\documentclass[conference]{IEEEtran}
\IEEEoverridecommandlockouts
\usepackage{cite}
\usepackage{amsmath,amssymb,amsfonts}
\usepackage{algorithmic}
\usepackage{graphicx}
\usepackage{textcomp}
\usepackage{xcolor}
\def\BibTeX{{\rm B\kern-.05em{\sc i\kern-.025em b}\kern-.08em
    T\kern-.1667em\lower.7ex\hbox{E}\kern-.125emX}}

\newsavebox\CBox

\newcommand{\cbit}{\begin{compactitem}}
	\newcommand{\ceit}{\end{compactitem}}
\newcommand{\cben}{\begin{compactenum}}
	\newcommand{\ceen}{\end{compactenum}}

\definecolor{OliveGreen}{rgb}{0,0.6,0}

\newcommand{\bit}{\begin{itemize}}
	\newcommand{\eit}{\end{itemize}}
\newcommand{\ben}{\begin{enumerate}}
	\newcommand{\een}{\end{enumerate}}
\newcommand{\beq}{\begin{equation}}
\newcommand{\eeq}{\end{equation}}

\newcommand{\hide}[1]{}

\usepackage{graphicx}
\usepackage{subfigure}
\usepackage{booktabs} 
\usepackage{paralist}
\usepackage{amssymb}

\usepackage{hyperref}
\usepackage{amsfonts} 
\usepackage{amsmath} 
\usepackage{bbm}
\usepackage{array,multirow,graphicx}
\usepackage{bm}
\usepackage{etoolbox,xspace}
\usepackage{footmisc}
\usepackage{algorithm}
\usepackage{algorithmic}  

\begin{document}

\title{Self-Supervision for Tackling Unsupervised Anomaly Detection: Pitfalls and Opportunities}

\author{
	\IEEEauthorblockN{Leman Akoglu}
\IEEEauthorblockA{\textit{Heinz College of Information Systems and Public Policy } \\
	\textit{Carnegie Mellon University}\\
	{lakoglu@andrew.cmu.edu}}
\and
\IEEEauthorblockN{Jaemin Yoo}
\IEEEauthorblockA{\textit{School of Electrical Engineering} \\
\textit{KAIST}\\
{jaemin@kaist.ac.kr}}
\and

}

\maketitle


\begin{abstract}
Self-supervised learning (SSL) is a growing torrent that has recently transformed machine learning and its many real world applications, by learning on massive amounts of unlabeled data via self-generated supervisory signals. 
Unsupervised anomaly detection (AD) has also capitalized on SSL, by self-generating pseudo-anomalies through  various data augmentation functions  or external data exposure. In this vision paper, we first underline the importance of the choice of SSL strategies on AD performance, by presenting evidences and studies from the AD literature.
Equipped with the understanding that SSL incurs various hyperparameters (HPs) to carefully tune,  
we present recent developments on unsupervised model selection and augmentation tuning for SSL-based AD. 
We then highlight emerging challenges and future opportunities; on designing new pretext tasks and augmentation functions for different data modalities, creating novel model selection solutions for systematically tuning the SSL HPs, as well as on capitalizing on the potential of pretrained foundation models on AD through effective density estimation.

\end{abstract}


\section{Introduction: Self-Supervised Learning for AD
}
\label{sec:intro}


Self-supervised learning (SSL) is a machine learning (ML) paradigm where the ML model trains itself to learn one part of the input data from another part.
SSL, which can learn from vast amounts of unlabeled data, is also called predictive or pretext learning as it transforms the unsupervised learning task into a supervised one by auto-generating the labels \cite{balestriero2023cookbook}.
It has been argued that SSL is likely  a key toward ``unlocking the dark matter of intelligence'' \cite{darkmatter21}, where Yann Lecun
has been one of the biggest advocates of SSL, at least as a means to making deep learning data-efficient, who stated ``\textit{If artificial intelligence is a cake, self-supervised learning is the bulk of the cake.}'' \cite{dickson2020}.
In fact, SSL has already started to take the world by storm, as embodied in large language models (LLMs) like OpenAI's ChatGPT and its many real world use cases \cite{openai2023gpt4}. 


SSL is particularly attractive for \textit{unsupervised} anomaly detection (AD) problems, for which acquiring labeled data is costly, laborious, in some cases impossible or even undesirable. To elaborate, it is hard in most settings to (pre)specify what constitutes anomalies, i.e. they are the ``unknown unknowns'', which makes labeling impractical. Anomalies also frequently appear in adversarial scenarios, and thus are subject to change rapidly. To stay alert to emerging threats, it is desirable to adopt unsupervised techniques, often in hybrid combination with supervised classifiers that have been trained on historical schemes \cite{bauder2017survey,soheily2018intrusion,carcillo2021combining}.   Therefore, in the absence of any labeled anomalies, SSL based techniques offer opportunities for many unsupervised AD problems in the real world. 

\begin{figure}
    \centering
    \includegraphics[width=0.4\textwidth]{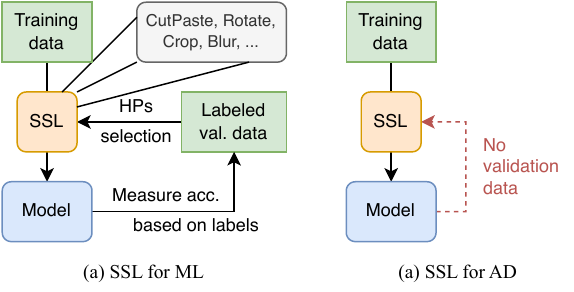}
    \caption{
        The main challenge in SSL for AD:  labeled validation data does \textbf{not} exist for tuning the  hyperparameters in SSL.
    }
    \label{fig:overview} 
\end{figure}

At the heart of SSL lies the pretext (or surrogate, self-supervised) task. Depending on the type of pretext learning, SSL methods have been organized into contrastive, predictive, and generative \cite{liu2021self,wu2021self}.
Contrastive methods typically employ data augmentation toward learning meaningful representations. Predictive methods create surrogate (or pseudo) labels from the data itself, often using masking strategies. Finally, generative methods aim to capture the underlying data distribution by trying to mimic the generative processes of the input data.


{
In this vision paper, we introduce recent developments of SSL for AD and essential challenges that have arisen from the literature.
We focus on the difficulty of augmentation tuning and model selection of SSL, given that a fair selection of HPs is infeasible in AD where no labeled data are given at training time for the purpose of validation.
Fig. \ref{fig:overview} illustrates why model selection is difficult especially on SSL for unsupervised AD.
}

{
We summarize the key take-aways as follows:
\begin{enumerate}
    \item SSL for AD is different from SSL for ML in essence, and it has the challenge of HP selection (Sec. \ref{sec:diff}).
    \item The choice of a pretext task is important for the success of SSL in general (Sec. \ref{sec:pretext}).
    Similarly, the choice of data augmentation plays a key role in SSL for AD (Sec. \ref{sec:aug}).
    \item We introduce recent works toward a fair and/or automatic selection of HPs for SSL for AD, focusing on the idea of transduction; leveraging unlabeled test data (Sec. \ref{sec:sol}).
    \item GenAI and foundation models can be the future of AD, provided massive amounts of training data exist (Sec. \ref{sec:gen}).
\end{enumerate}
}

\section{SSL for ML vs. AD: A Key Difference}
\label{sec:diff}


There exists a key difference on the purpose of using SSL in the traditional ML literature versus the AD literature, which is respectively, \textit{generalization} vs. \textit{pseudo-anomaly generation}. 
The use of SSL in ML toward better generalization is akin to complementing the sparsely-sampled true data manifold via ``filling in'' the space with more \textit{positive} samples; for example, mirror-image of a dog or a dog wearing a raincoat is still a dog. In contrast, employing SSL in AD toward pseudo-anomaly generation is akin to ``filling in'' the inlier-only input space with \textit{negative} samples; for example, various augmentations are employed in \cite{xu2022calibrated} to learn a better one-class (inliers) boundary than one could learn with unsupervised (deep) SVDD alone \cite{tax2004support,ruff2018deep}. 
Typically the pseudo-anomalies are generated in one of two ways: ($i$) via data augmentation \cite{Golan18GEOM,Li21CutPaste} or 
 ($ii$) via external data  for outlier exposure \cite{Hendrycks19OE,liznerski2022exposing}.


While perhaps re-branding under the name SSL, the idea of injecting artificial anomalies to inlier data to create a labeled training set for AD dates back to the early 2000s \cite{theiler2003resampling,steinwart2005classification,abe2006outlier}.
Fundamentally, under the uninformative/uniform prior for the (unknown) anomaly-generating distribution, these methods are asymptotically consistent density level set estimators for the support of the inlier data distribution \cite{steinwart2005classification}. Unfortunately, they are ineffective and sample-inefficient in high dimensions (such as for image data) as they require a massive number of sampled anomalies to properly ``fill in'' the sample space.


With today's SSL methods for AD, we see a shift toward various 
 \textit{non-uniform} priors on the distribution of anomalies. 
In fact, current literature on SSL-based AD is laden with many forms of generating pseudo-anomalies (numerous different augmentations, and potentially infinitely many external ``exposure'' datasets), each introducing its own inductive bias.
As a consequence, success on a given AD task depends on \textit{which} augmentation function is used or \textit{which} external dataset the learning is exposed to as pseudo anomalies, and importantly ``to what extent the pseudo anomalies mimic the nature of the true (yet unknown) anomalies'' in the test data \cite{yoo2023data}.

Not surprisingly, there exist evidences in the literature that the choice has a significant impact on the outcome.
For example, Golan \textit{et al.} \cite{Golan18GEOM} have shown that geometric transformations create better pseudo-anomalies than pixel-wise augmentations for detecting semantic class anomalies.
In contrast, Li \textit{et al.} \cite{Li21CutPaste} have reported that (global) geometric transformations \cite{Golan18GEOM} fail at detecting small defects in industrial object images, where (local) augmentations such as random cut-and-paste perform significantly better.
In their eye-opening study, Ye \textit{et al.} \cite{Ye21Bias} have observed that sampling pseudo-anomalies from a biased subset of true anomalies
leads to a biased error distribution; 
the test error is lower on the seen type of anomalies during SSL training, at the expense of much larger error on unseen anomalies---even when the unseen anomalies are easily detected by an \textit{unsupervised} detector (!).
Most recently, we have confirmed and replicated these findings in several other experimental settings \cite{yoo2023data}.

\section{Role of The Pretext Task}
\label{sec:pretext}


A key question for SSL is: what pretext task would be most useful to various downstream tasks of interest. 
Perhaps the most successful and groundbreaking use of SSL has been in large language models (LLMs) like OpenAI's ChatGPT \cite{openai2023gpt4}, where the pretext task is of the ``fill in the blanks'' nature, like predicting the arbitrarily masked words or predicting the next sentence in human-generated text. 
One would most likely agree that being able to predict the last few pages of a mystery novel indeed would be demonstrative of solid reading comprehension and text understanding.\footnote{This example is given by Ilya Sutskever (Co-Founder and Chief Scientist, OpenAI) during an interview by Alexandr Wang (CEO and Founder, Scale AI), available online at \url{https://www.youtube.com/watch?v=UHSkjro-VbE}.} 
However, it is not obvious how it extends to other data modalities, in other words, what it means to ``understand'' images, videos, time series, etc.

Another perspective on the challenge of going beyond natural language to other data modalities such as images and designing  pretext tasks analogous to ``filling in the missing parts'' is that they are high dimensional continuous objects, as opposed to a finite set of discrete words, over which it is yet unknown how to represent suitable probability distributions \cite{darkmatter21}. 
There are an infinite number of possible missing image patches, video frames or speech segments. Representing all possible high-dimensional continuous outcomes with suitable probability distributions seems like an intractable problem.


Masking, or the ``fill in the blanks'' style pretext learning,  may not be universally applicable to all data modalities. 
It may also not be suitable for all downstream tasks. 
Even though communities other than the NLP community have also used masking strategies to design pretext tasks for images and videos 
in computer vision \cite{he2022masked,tong2022videomae}, the majority of downstream evaluations has been limited to image classification tasks, with less emphasis on other downstream tasks such as segmentation, object counting, object detection, etc.  \cite{bordes2022high} 
Others have also studied the 
interplay between generalization, augmentation and inductive bias induced by SSL \cite{cabannes2023ssl}.


A recent paper by Balestriero and Lecun  \cite{balestriero2022contrastive} has shown that the surrogate/pretext task is to help solve the supervised downstream task to the extent that two similarity matrices -- namely, one dictated by the pretext task and the other corresponding to the downstream labels -- have certain matching spectral properties, which highlights the important concept of pretext--downstream task alignment. 
In a recent comprehensive study on SSL-based AD, we have observed similar alignment phenomena, where 
we find that the AD performance benefits from self-supervision to the extent that the pseudo-anomaly generation is capable of mimicking the true anomalies in the test data, which otherwise can even impair performance \cite{yoo2023data}. We elaborate on our findings in the following section. 


In summary, it remains an open problem to choose a pretext task for SSL that guarantees good generalization to a suite of downstream tasks for different data modalities. As Yann Lecun stated, 
``\textit{The big question is: can you build those surrogate tasks without requiring expensive manual labeling [$\ldots$] and drive the system to learn the \textbf{right} things}.''\footnote{Yann Lecun interview at NeurIPS 2022, available online at  \url{https://www.youtube.com/watch?v=9dLd6n9yT8U}.} 

\section{Data Augmentation Matters}
\label{sec:aug}

In many applications of SSL to unsupervised anomaly detection, an augmentation function is applied to the inlier samples to synthesize or self-generate anomalous examples. For image data, for instance, there exists a plethora of augmentation functions; such as rotation, blurring, cropping, masking, color inverting, to name just a few. 
These different functions, as a result, generate pseudo anomalies of different nature. 

The literature has reported multiple evidences that the choice of the augmentation function has key impact on detection performance. For instance, geometric augmentations have been the choice for semantic anomalies \cite{Golan18GEOM}, where the inliers and true anomalies are images from different classes. In stark contrast, these global augmentations, such as rotation or flipping, fail substantially in detecting small industrial defect anomalies as reported by Li {\em et al.} \cite{Li21CutPaste}---it is exactly why they proposed new, local augmentations for defect detection that involve small perturbations such as cutting and pasting of small image patches. 
In another recent work, Ding {\em et al.} explored five different augmentations and applied CutMix \cite{yun2019cutmix} on a subset of their datasets, whereas on the remaining, medical datasets, they chose to use samples from an external (also medical) dataset for outlier exposure. 

More examples can be listed from the SSL-based AD literature, wherein different augmentation choices are made depending on the dataset/task. 
Besides the discrete choice of which augmentation function, 
one also has to choose
associated continuous HPs (e.g. width and height of patch size to cut-out, rotation degree, etc.).
The issue in the literature is that such choices are either not justified or made in an after-the-fact manner. One could intuitively imagine that local augmentations like cut-paste would be more suitable for small industrial defect-style anomalies as compared to gross augmentations like flipping an entire image. Similarly, one could agree that outlier-exposing a medical dataset from another medical dataset would be more suitable than any other arbitrary dataset. Yet, given the infinite pool of such choices, SSL-based AD community does not systematically recognize data augmentation as a hyperparameter (HP) \cite{yoo2023data}. This is in contrast to the supervised SSL literature, which explicitly tunes data augmentation as a HP \cite{mackay2019self,zoph2020learning,ottoni2023tuning,Cubuk19Auto}.
The situation has come to a point akin to the issue of ``p-value hacking'' \cite{ziliak2017p} in statistics-related fields, i.e. SSL-based AD 
has become a playground for what-we-call ``augmentation snooping/fishing''.


The ``fishing'' issue is not limited to data augmentation (hyperparameters) and SSL-based AD specifically, but goes beyond more broadly to general AD model/hyperparameter (HP) selection at large. As we showed in recent work \cite{ding2022hyperparameter}, alarmingly, the reported performance results in recent publications on deep AD models are systematically higher than one would obtain by random picking, i.e. average/expected performance across HP choices, in the absence of any other knowledge (or sneak-peek (!)) of ``good'' HP values.
While the proper configuration of HPs is critical to performance outcomes and both shallow and especially deep AD models with a longer list of HPs are sensitive to HPs, the AD community seems to have turned a blind eye to the issue, rendering (SSL-based) AD model selection ``the elephant in the room''---a major problem that is obviously present but avoided as a subject for discussion because it is more comfortable to do so. In fact, it is not uncommon to find deep AD models in the literature where criteria/justification for the ``author-suggested HPs'' are swept under the rug. Some work even  report results based on HPs that are tuned on the \textit{test} data (!), e.g. \cite{zhou2017anomaly,ruff2018deep,akcay2019ganomaly}, violating fundamental ML principles and practices. 

Admittedly, systematically tuning HPs (i.e. model selection) is nontrivial for unsupervised settings, in the absence of any labeled validation/hold-out data \cite{ma2023need}, although, the AD literature has been  growing recently with novel ideas on unsupervised outlier model selection (UOMS) \cite{MetaOD21,zhao2022toward,zhao2022towards,hyper23}. 
This vision paper is another effort toward drawing the community's attention to this fundamental problem, which we coined as UOMS. In the following, we present our recent work on unsupervised augmentation tuning for SSL-based AD specifically. 

\section{Toward Self-tuning SSL-based AD}
\label{sec:sol}

In the supervised setting, various models with different HP choices are trained and then evaluated on hold-out validation data, which is labeled. This provides an estimate of the generalization performance of each model and is used to select the best HP configuration. 
In unsupervised settings, such labeled validation data does not exist. This makes UOMS relatively a much more difficult problem. 

Earlier efforts toward UOMS have proposed \textit{internal} evaluation measures. In principle, they quantify various properties of the outlier scores (e.g. bi-modality, clusteredness, etc.) \cite{MarquesCZS15,journals/corr/Goix16,JCC8455}, model parameters (esp. for NN-based deep models) \cite{martin2021predicting,yang2023test}, as well as consensus among the trained models \cite{lin2020infogan,DuanMSWBLH20} to deduce which models are likely to have detected the anomalies.
Unfortunately, however, in an extensive measurement study we have found those internal measures to be insufficient \cite{ma2023need}---most being statistically indifferent from random choice or unable to outperform basic ensemble models like IsolationForest \cite{Liu08IF} with default HPs. 

The key challenge for UOMS is designing an effective unsupervised validation loss. While the former internal measures can be used to this end for SSL-based AD models as well, they are neither effective nor specific to SSL-based AD. In recent work, we have capitalized on the specifics of data augmentation to design novel validation losses for SSL-based AD \cite{Yoo23DSV}. The key idea is to quantify the \textit{alignment} between the inlier data that is augmented with pseudo anomalies, i.e. $\mathcal{D}_{\text{in}} \cup \mathcal{D}_{\text{aug}}$ and the given (unlabeled) test data (containing both inliers and true anomalies), i.e. $\mathcal{D}_{\text{test}}$. The working assumption is that SSL-based AD can be effective to the extent that the pseudo anomalies mimic the true anomalies well. The alignment can be measured in the input space as well as the embedding space for NN-based models.  For example, such an alignment loss aims to quantify if cut-paste augmented samples are better aligned with or similar to true samples with industrial defects than are rotated samples, so as to deduce cut-paste to be more suitable than rotation augmentation. 

Notice that such an alignment can be measured only in the presence of the test data $\mathcal{D}_{\text{test}}$. 
In essence, 
the key concept we leverage is \textit{transduction} or trunductive learning, as advocated by Vladimir Vapnik
who stated ``\textit{When solving a problem of interest, do not solve a more general problem as an intermediate step. Try to get the answer that you really need but not a more general one.}'' \cite{vapnik2006estimation}. The principle is  to \textbf{not} try to induce from having solved an  intermediate problem (in this case, estimating a general decision boundary between inliers and all potential anomalies) that is no simpler or is more general/complicated/involved than the original problem at hand (detecting the specific, observed anomalies in test data).
We remark that using the test data for model selection in this scenario does not violate the fundamental ML principle of ``no training on test data'', since importantly, the test data here is \textit{unlabeled}.  This is simply the transductive learning setting, where test data is given during training. In fact, most AD tasks occur under this scenario where a bulk of unlabeled data is provided in which anomalies are to be identified (e.g. a database of CT scans, medical claims, transactions,  etc.).

Provided with an unsupervised validation loss, one can employ various HP optimization/search techniques such as the 
grid search, random search, SMBO, and others \cite{Bischl23Survey}.
What is more, one can leverage gradient-based techniques as long as both the validation loss and the augmentation  can be written as differentiable functions. Our recent attempt in this direction introduced the first end-to-end augmentation tuning framework for SSL-based AD, and utilized a differentiable, unsupervised alignment validation loss along with differentiable analytical formulas for various augmentation function choices 
\cite{Yoo23End}.

The key principle here is to recognize various choices within SSL as HPs and aim to systematically tune those choices to achieve robust detection performance on any input task.



\section{GenAI's Potential for AD}
\label{sec:gen}

In Sec. \ref{sec:diff} we discussed a key difference between SSL for ML vs. SSL for AD as it relates to the usage purpose of data augmentation (respectively, generalization vs. pseudo anomaly synthesis).
At the same time, there also exists a common thread to both supervised ML and unsupervised AD---that is, to learn the underlying data manifold as effectively as possible given finite training data. If one could successfully capture the inlier data distribution, then, anomalies could be detected effectively as low probability/likelihood instances. 

On one hand, this indirect problem (i.e. density estimation) appears to be a more general/complicated/involved problem than the one at hand (i.e. anomaly detection). As such, recalling Vapnik's statement that we quoted in Sec. \ref{sec:sol}, density estimation may not appear as the most straightforward route to anomaly detection. 
In fact, it has been argued that ``\textit{we may never have techniques to represent suitable probability distributions over high-dimensional continuous spaces [as with all possible image patches or video frames]}'', and that it ``\textit{seems like an intractable problem}'' \cite{darkmatter21}.

On the other hand, today's generative models are taking the world by storm \cite{zhang2023complete}, achieving outstanding results in learning data distributions by capitalizing on ($i$) massive amounts of (pre)training data, ($ii$) large-scale compute power and ($iii$) highly expressive, billion-scale parameterized transformer models. 
Today's mostly autoregressive or diffusion based generative models are able to learn the underlying data distribution sufficiently well from massive amounts of unlabeled data (in other words, very densely sampled data manifold), to the extent that they can generate realistic, human-like content like conversations \cite{openai2023gpt4} and images \cite{ramesh2021zero}.
As such, the opportunities that GenAI offers and its potential impact on anomaly detection should not be overlooked, since density estimation/pattern mining and anomaly detection are interlinked problems, i.e. two sides of the same coin \cite{schubert2015outlier}. 

The current bottlenecks for the advancement of AD via generative models seem to lie on the necessity for massive (in this case, inlier) data as well as immense compute sources. The former is particularly challenging for domains to which AD applies, where the amount of data is limited, proprietary, or otherwise costly to obtain (e.g., wet-lab experiments, accounting data, medical imaging, etc.). In the future, democratizing such large-scale pre-trained models on a broader range of data modalities, beyond text and images, has the potential to break new ground for AD in various domains.


\section{Summary: Take-aways and Future Research}
\label{sec:conclusion}

Self supervised learning (SSL) has been transformative in many applications of ML in the real world. Self-generation of supervisory signals has particularly attracted the unsupervised anomaly detection (AD) literature. Through this article, we underlined the importance of the pretext task for SSL-based AD. Simply put, we highlighted that the choice of augmentation or the external exposure dataset strongly impacts detection performance depending on the input AD task. 
This suggests that SSL hyperparameters (HPs) should be tuned systematically for robust outcomes. 
To this end, we summarized recent work on UOMS (unsupervised outlier model selection) at large, as well as transductive augmentation tuning specific to SSL-based AD.

AD applies to numerous diverse domains, such as manufacturing, finance, medicine, security, surveillance, and so on, all of which exhibit multi-modal data. While pretext tasks for text and images are plenty, future work can investigate which pretext tasks would be best suited to other data modalities such as tabular data or (multivariate) time series.
Further work is needed on new augmentation functions for complex data modalities, which can flexibility mimic a large variety of anomaly types; e.g. spikes,  motif shifts, trend changes, etc. in time series and conditional, collective, global outliers, etc. in tabular data.
AD community should also keep a keen eye on the potential of  foundation models in breaking new ground for AD as effective density estimators/manifold learners.






\bibliography{00ref.bib}
\bibliographystyle{IEEEtran}
\end{document}